\documentclass[letterpaper, 10 pt, conference]{ieeeconf}
\usepackage{array}
\usepackage{textcomp}
\usepackage{stfloats}
\usepackage{subfigure}
\usepackage{url}
\usepackage{verbatim}
\usepackage{graphicx}
\usepackage{cite}
\usepackage{hyperref}

\usepackage{amsmath, amsfonts, amssymb, amsthm}
\usepackage{mathtools}
\usepackage{ulem}

\IEEEoverridecommandlockouts
\hypersetup{hidelinks}
\begin{document}
\title{\LARGE \bf Gate-Aware Online Planning for Two-Player Autonomous Drone Racing}
\author{Fangguo Zhao$^{1}$, Jiahao Mei$^{2}$, Jin Zhou$^{1}$,  Yuanyi Chen$^{1}$, Jiming Chen$^{1}$, and Shuo Li$^{1}$
\thanks{$^{1}$Authors are with the College of Control Science and Engineering, Zhejiang University, Hangzhou 310027, China {\tt\small shuo.li@zju.edu.cn}}%
\thanks{$^{2}$Jiahao Mei is with the Department of Automation, Zhejiang University of Technology, Hangzhou 310023, China.}
}

\maketitle
\begin{abstract}
The flying speed of autonomous quadrotors has increased significantly over the past $5$ years, particularly in the field of autonomous drone racing. However, most research primarily focuses on the aggressive flight of a single quadrotor, simplifying the racing gate traversal problem to a waypoint passing problem that neglects the orientations of the racing gates. In this paper, we propose a systematic method called Pairwise Model Predictive Control (PMPC) that can guide two quadrotors online to navigate racing gates with minimal time and without collisions. The flight task is initially simplified as a point-mass model waypoint passing problem to provide analytical time optimal reference through an efficient two-step velocity search method. Subsequently, we utilize the spatial configuration of the racing track to compute the optimal heading at each gate, maximizing the visibility of subsequent gates for the quadrotors. To address varying gate orientations, we introduce a novel Magnetic Induction Line-based spatial curve to guide the quadrotors through racing gates of different orientations. Furthermore, we formulate a nonlinear optimization problem that uses the point-mass trajectory as initial values and references to enhance solving efficiency, enabling the method to run onboard at a frequency of $200$ Hz. The feasibility of the proposed method is validated through both simulation and real-world experiments. In real-world tests, the two quadrotors achieved a top speed of $6.1 m/s$ on a $7$-waypoint racing track within a compact flying arena of $5 m \times 4 m \times 2 m$.

\end{abstract}
\section*{Supplementary Material}
\href{https://www.youtube.com/playlist?list=PLYVbdwXarwXDDZg1oW_AbgKfCN5pgoRDZ}{Video: \url{https://www.youtube.com/playlist?list=PLYVbdwXarwXDDZg1oW_AbgKfCN5pgoRDZ}}
\section{Introduction}
  
In recent years, autonomous quadrotors have consistently broken speed records and demonstrated numerous potential applications, such as emergency delivery services, search and rescue, and rapid exploration in unknown areas. One of the driving forces behind these new records is autonomous drone racing, a platform for testing various elements of a drone's autonomous agile flight, including navigation, trajectory generation, and control techniques \cite{10530312}.
However, the majority of the research on this topic mainly focuses on the aggressive flight of a single drone. Multi-drones with aggressive flight maneuvers can significantly improve the efficiency of flying tasks\cite{8424838}. In this work, we focus on this problem and propose a method that can generate time-optimal trajectories for two quadrotors online, allowing them to fly through gates of different orientations without collisions.

Many research achievements have been made in trajectory generation for high-speed flights of single quadrotors. A classic aggressive trajectory generation method is based on differential flatness and minimum-snap techniques \cite{mellinger2011minimum, faessler2017differential}, with subsequent improvements \cite{zhou2021ego, 9543598, 10342456, fork2023euclidean, qin2024timeoptimalplanninglongrangequadrotor} aimed at enhancing computational efficiency and safety for quadrotors. However, these trajectories cannot guarantee time optimality due to their inherent smoothness \cite{10530312}. Recently, a milestone work on polynomial trajectory optimization called MINCO \cite{WANG2022GCOPTER} has been developed to handle multiple constraints during flight. Some works have used MINCO to directly optimize flight time, achieving near time-optimal flight \cite{10610148}. Another approach converts time optimal trajectory generation into a nonlinear optimization problem, using complementary progress constraints (CPC) to ensure the quadrotor passes waypoints in sequence \cite{foehn2021time}. While this method produces globally optimal trajectories, it often requires dozens of hours to solve the optimization problem. For online aggressive flight, Model Predictive Contouring Control (MPCC) is a promising method that maximizes progress along the reference trajectory and minimizes tracking errors integrating point-mass trajectory generation \cite{romero2022model}. Recently, with the development of deep learning techniques, there have been advancements in achieving high-speed agile flight through imitation learning \cite{r6,ferede2024end,zhou2024imitation} and reinforcement learning methods \cite{song2023reaching, kaufmann2023champion}. However, the former requires a pre-established large trajectory library, which still cannot cover the full state spaces of quadrotors, while the latter is only effective for specific tracks, limiting its application in the real world.

\begin{figure}[t]
    \centering
    \includegraphics[width=0.45\textwidth,trim={0cm, 0cm, 0cm, 0cm}, clip]{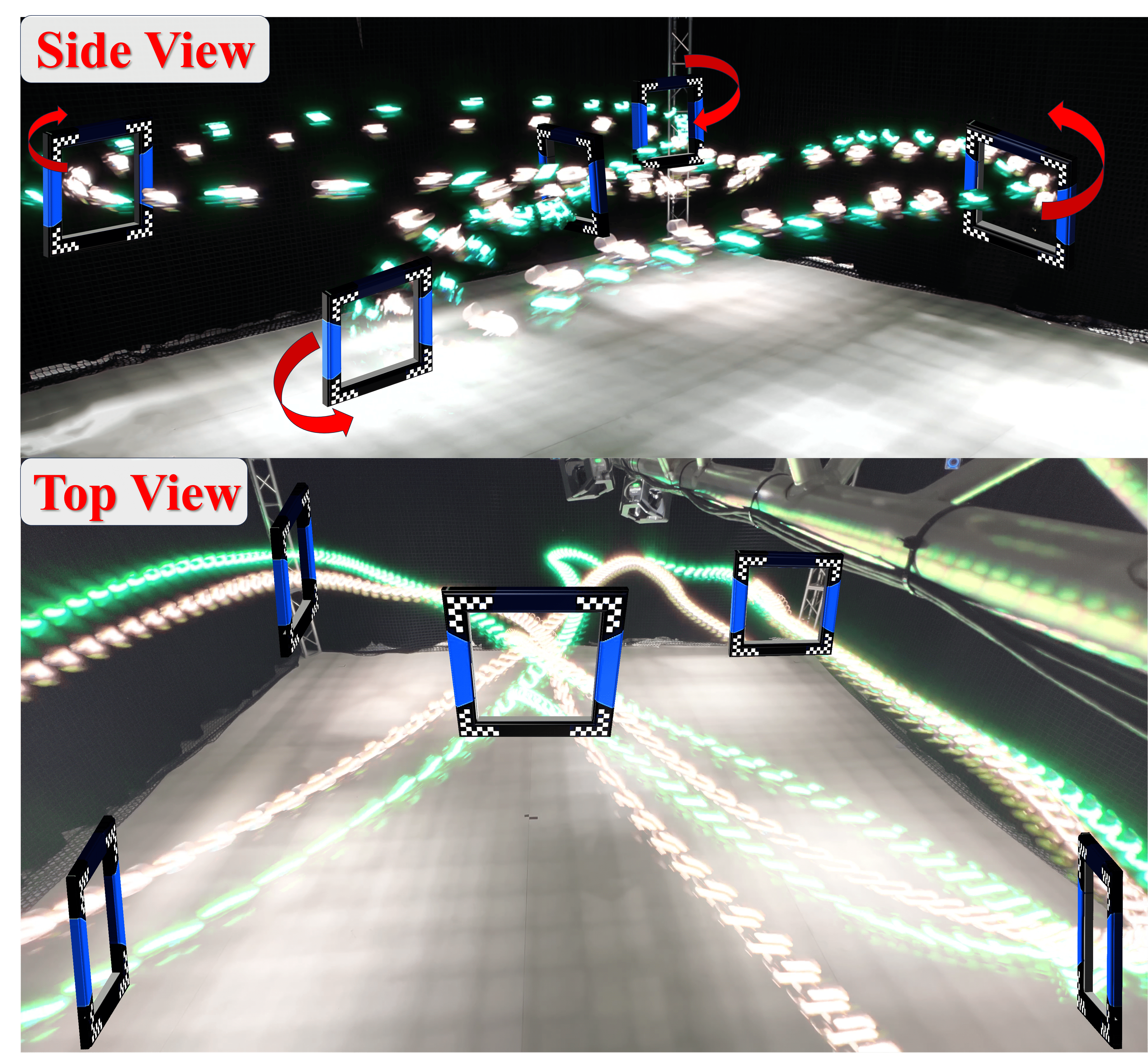}
    \vspace{-1em}
    \caption{Two quadrotors race in an $8$-shape track with two different gate orientations and achieve a top speed of $6.1m/s$.}
    \label{fig:two quadrotor racing photo}
    \vspace{-1em}
\end{figure}

Previous works simplify the racing gate to a fixed center point, which does not consider the passing direction of the gate in the optimization problems \cite{foehn2021time, romero2022model}. However, different tracks contain gates with various orientations, making it necessary to consider the direction from which the drone will fly through the gate to prevent collision with the gates. Several studies construct a flight corridor along the trajectory to limit the flight space \cite{krinner2024mpccmodelpredictivecontouring,10610148,10610179,9543598,10.1007/978-3-030-71151-1_4}. Some methods convert the drone dynamics into the Frenet-Serret frame \cite{9811764} to address this issue. Alternatively, reinforcement learning (RL) methods employ local safety rewards \cite{song2021autonomous} to encourage the drone to pass near the middle of the four gate corners, implicitly considering the gate orientation \cite{geles2024demonstratingagileflightpixels}. However, the flight corridor restricts the flight space, and the corner constraints do not account for how the direction of passage might be influenced by the adjacent waypoints.
In this work, we address this problem by proposing a novel Magnetic Induction Line (MIL)-based curve to connect each gate as a spatial reference when approaching the gate.

Swarms of quadrotors have been extensively studied for formation flight \cite{10219410}, search and rescue \cite{mcguire2019minimal}, and exploration tasks \cite{8424838}. However, time-optimality or aggressive flight is not the focus of these studies. In these works, collision avoidance within a swarm is typically addressed in an intuitive manner, where others' trajectories are treated as static constraints during optimization. This method requires check and re-check after optimization to avoid the ego-optimized trajectory from colliding with others' new trajectories known as \textit{deconfliction} \cite{tordesillas2021mader, 10161244}. The complex procedure is not suited for time-critical tasks, such as drone racing. Game theory has also been applied to autonomous drone racing with multiple-quadrotors. However, the flight speeds achieved in the research are still far from the quadrotors' limits \cite{spica2020real,di2023cooperative}. For high-speed multi-quadrotor flight tasks, existing work uses nonlinear optimization methods to generate optimal trajectories \cite{10341844, imav2023-31:y_shen_et_al}. However, this approach requires hours of offline computation.

\begin{figure*}[ht]
    \centering
    \includegraphics[scale=0.23, trim={6cm, 29cm,  0cm, 0cm},clip]{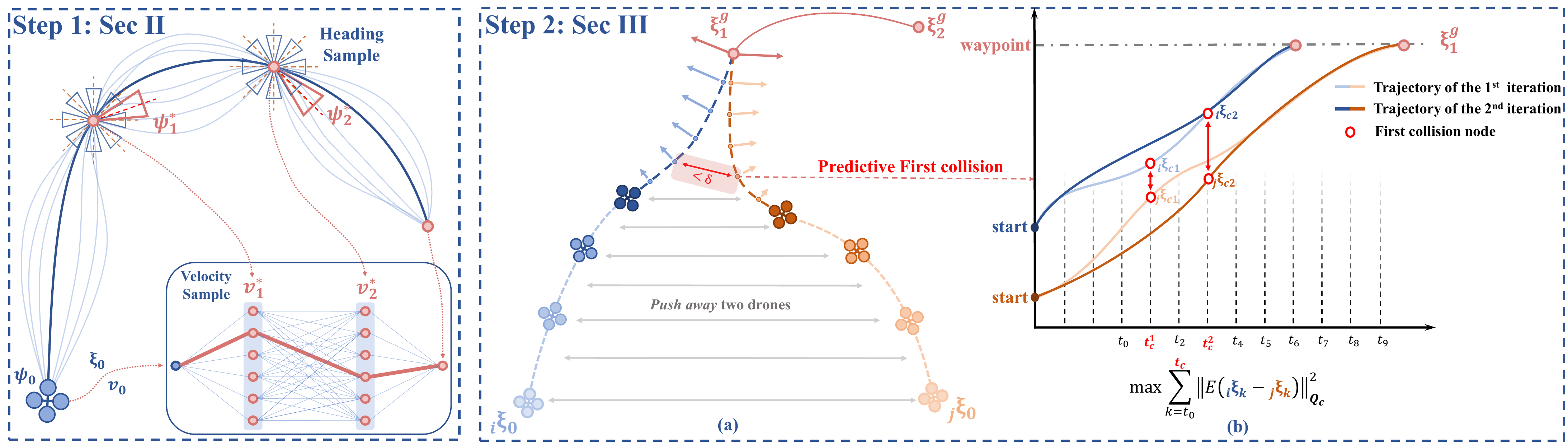}
    \vspace{-1em}
    \caption{The pipeline of the proposed method. Step 1: A point-mass model is used to generate the global time-optimal trajectory which is used as initial values for the optimization. Step 2: An optimization problem is established to guide two quadrotors to fly through the waypoints without collision.}
    \label{fig:main_figure}
    \vspace{-1em}
\end{figure*}

In this paper, we propose a systematic method for generating time-optimal trajectories online for two-player racing drones, enabling them to pass through waypoints with varying orientations in a predefined sequence (Fig. \ref{fig:main_figure}). The contributions of our study are as follows:
\begin{enumerate}
\item An augmented dynamics MPC called Pairwise-MPC is developed to guide two quadrotors to fly through gates online without collision.
\item An efficient point-mass and heading trajectory generation method that maximizes next-gate visibility with minimum time.
\item A novel magnetic field-inspired curve serves as a spatial reference to guide drones to fly through gates at different orientations.
\item The method is verified in both simulation and real-world experiments, where the quadrotors can achieve a top speed of $6.1m/s$ in a compact flight space.   
\end{enumerate}

\section{Point-mass Reference Trajectory and Magnetic Induction Line for Gate Traversing }
\label{sec:gtg}
Inspired by \cite{romero2022model}, we first treat the gate as a waypoint and simplify the quadrotor to a point-mass model, which we use to analytically generate time-optimal trajectories. Next, we determine the optimal yaw angles at each waypoint by considering the perception of the racing gates. These desired yaw angles are then connected using polynomials based on the minimum snap principle. The resulting position and heading reference trajectories are referred to as the \textit{point-mass reference trajectory}. Additionally, we design a Magnetic Induction Line (MIL) to guide the quadrotor to fly vertically through the racing gates. The point-mass reference trajectory and the MILs serve as references for the quadrotors, which will be explained in detail in Sec \ref{sec:PMPC}.

\subsection{Time-Optimal Point-Mass Reference Trajectories}
% \label{sec:traj}
For a three-dimensional point-mass dynamics model, the solution of the time-optimal trajectories passing waypoint sequentially is of bang-bang type. The trajectory can be generated by constructing a search graph $\mathcal{G} = (\mathcal{V}, \mathcal{E})$ as shown in Fig. \ref{fig:main_figure} (Step 1) where the nodes $\mathcal{V}$ are the velocities at each waypoint and the edges ${\mathcal{E}}$ are the travel time $T$ between the waypoints. Then the Dijkstra method can be used to find the optimal velocities in the graph.
For the details of getting the trajectory, we refer readers to \cite{romero2022time}.

\begin{figure}[hbt]
    \centering
    \includegraphics[width=0.49\textwidth,trim={0 0cm 0cm 0}, clip]{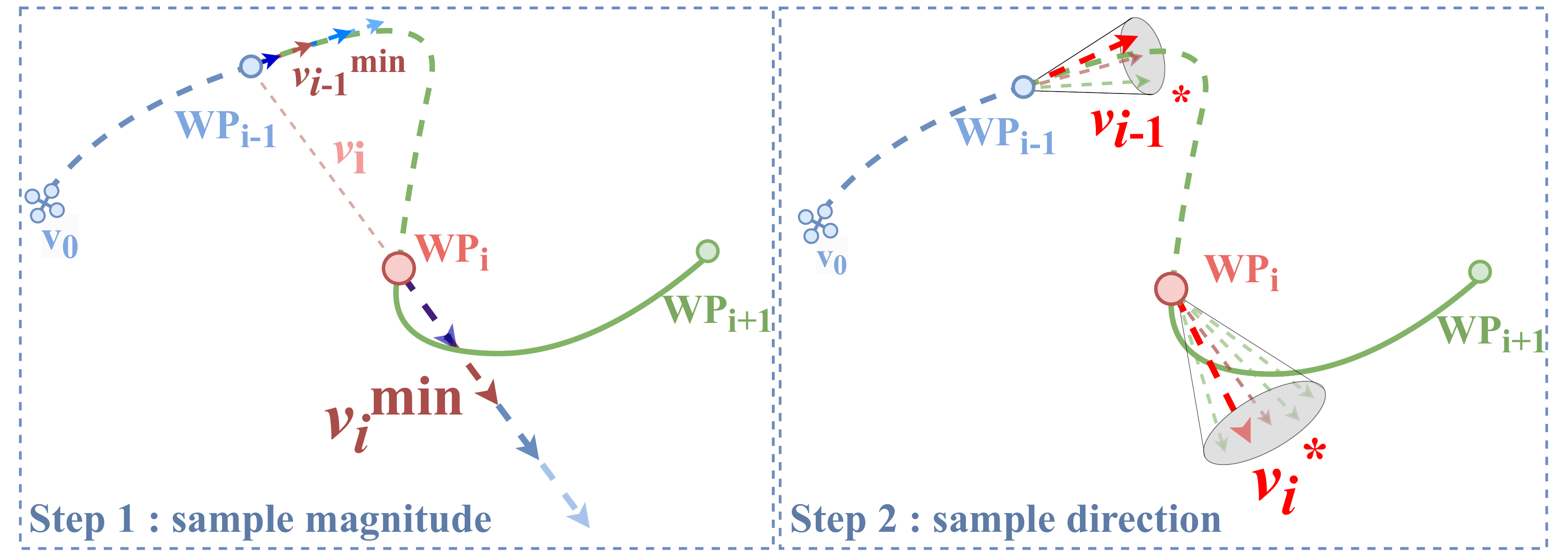}
    \caption{Step 1: sample velocities in the direction of $\mathbf{v}_i$ but in different magnitudes. Find the best velocities $v_i^{\rm min}$. Step 2: resample the velocities around $v_i^{\rm min}$ in different directions. Find the optimal velocity $v_i^{*}$.   }
    \label{fig:two_step}
    \vspace{-1em}
\end{figure}

Different to \cite{romero2022time}, to further improve the searching efficiency, we propose a two-step sampling method to sample the velocities and construct the graph without iterative velocity sampling as shown in Fig. \ref{fig:two_step}. 
We first fix the direction of the velocities $\mathbf{v}_i$ at waypoint $i$ by pointing from waypoint $\boldsymbol{\xi}_{i-1}$ to waypoint $\boldsymbol{\xi}_{i}$ and sample the magnitude of the velocities. 
\begin{align}
 % \frac{}}{}.
\mathbf{v}_i = {\boldsymbol{\xi}_{i} - \boldsymbol{\xi}_{i-1}}/{\left\|\boldsymbol{\xi}_{i} - \boldsymbol{\xi}_{i-1}\right\|_2}.
\label{eq:direction}
\end{align}
For each waypoint, we generate $20$ velocity samples by 
\begin{align}
\mathcal{V}_i=\{1 \times \mathbf{v}_i, 2 \times \mathbf{v}_i, ..., 20 \times \mathbf{v}_i \}.
\label{eq:candidates at waypoint}
\end{align}

Then, the best velocities $v_i^{\rm min}$ at each waypoint can be found by the Dijkstra method.
To further optimize the velocities at the waypoints, the second step is to regenerate $20$ velocity samples with the same magnitude with $v_i^{\min}$ but in different directions by aligning a predefined cone to $v_i^{\min}$ (Fig. \ref{fig:two_step} Step 2). Then a new graph $\mathcal{G}^{*}$ is constructed and we conduct the graph search again to find the optimal velocity $v_i^{*}$ at each waypoint. The point-mass trajectory can be generated by the method introduced in \cite{romero2022time} as a receding horizon approach.

However, the point-mass trajectory mentioned above only considers time-optimal waypoint passing and does not account for the perception of racing gates during flight. To address this limitation, we seek to determine the optimal heading angle $\psi_i^*$ at each waypoint, which maximizes the visibility of subsequent gates. Rather than simply orienting towards the next gate, our approach also considers the perception of all subsequent gates in the sequence. Therefore, the optimal heading angle $\psi_i^*$ is determined by solving the following optimization problem:
\begin{equation}
    \underset{\psi_i}{\min}\quad\mathcal{L}=\sum_{k=i}^{N_{wp} - 1}\gamma^{(k-i)} \left\|\boldsymbol{R}(\psi_i)(\prescript{}{w}{\boldsymbol{\xi}}_{k+1}^g -\prescript{}{w}{\boldsymbol{\xi}}_{i}^g)\right\|_2^2,
    \label{eq:yaw_optimal}
\end{equation}
where $\prescript{}{w}{\boldsymbol{\xi}}_i^g$ denote the positions of $i^{th}$ gate in world frame, $0<\gamma<1$ is the weighting term for all the coming gates, the farther the gate is, the less weight it will have. $\boldsymbol{R}(\psi_i)$ is rotation matrix only determined by $\psi_i$.
Solving problem (\ref{eq:yaw_optimal}) at each waypoint gets the optimal heading sequence. Due to the lower control effectiveness of heading control \cite{9794477}, we prefer smooth yaw maneuvers. Therefore, we use minimum angular accelerator based polynomial trajectory to connect the adjacent yaw angles $\psi^*_i$ and $\psi^*_{i+1}$. The coefficients of the polynomials can be computed efficiently using qpOASES\footnote{https://github.com/coin-or/qpOASES} \cite{mellinger2011minimum} with the time of the point-mass trajectory $T^*_i$. With this heading reference trajectory generation strategy, the drone will start to point to the next gate $i+1$ at gate $i$, which will be beneficial for the perception of the next gates and also is similar to how human pilots steer their drones\cite{pfeiffer2021human, 10610179}.
\subsection{Magnetic Induction Line for Gate Traversing}
\label{sec:mag_field}
The trajectory generation method described above treats the racing gate as a simple waypoint, without addressing the direction from which the quadrotor approaches the gate. However, in drone racing scenarios, quadrotors are expected to fly vertically through the racing gates. Therefore, we need to modify the designed point-mass trajectories to ensure that the quadrotor is guided through the gates in predefined directions, especially when it is close to the gates. Our goal is to find a curve that (a) connects two gates, (b) is perpendicular to the gates at both the start and end points, and (c) does not deviate significantly from the point-mass trajectories in the middle. This curve does not influence the flight when the quadrotor is in the middle of the two gates, but as the quadrotor approaches a gate, the curve will attract it to fly through the gate vertically.

\begin{figure}[hbt]
    \centering
    \includegraphics[scale=0.1,trim={0 0cm 0cm 0}, clip]{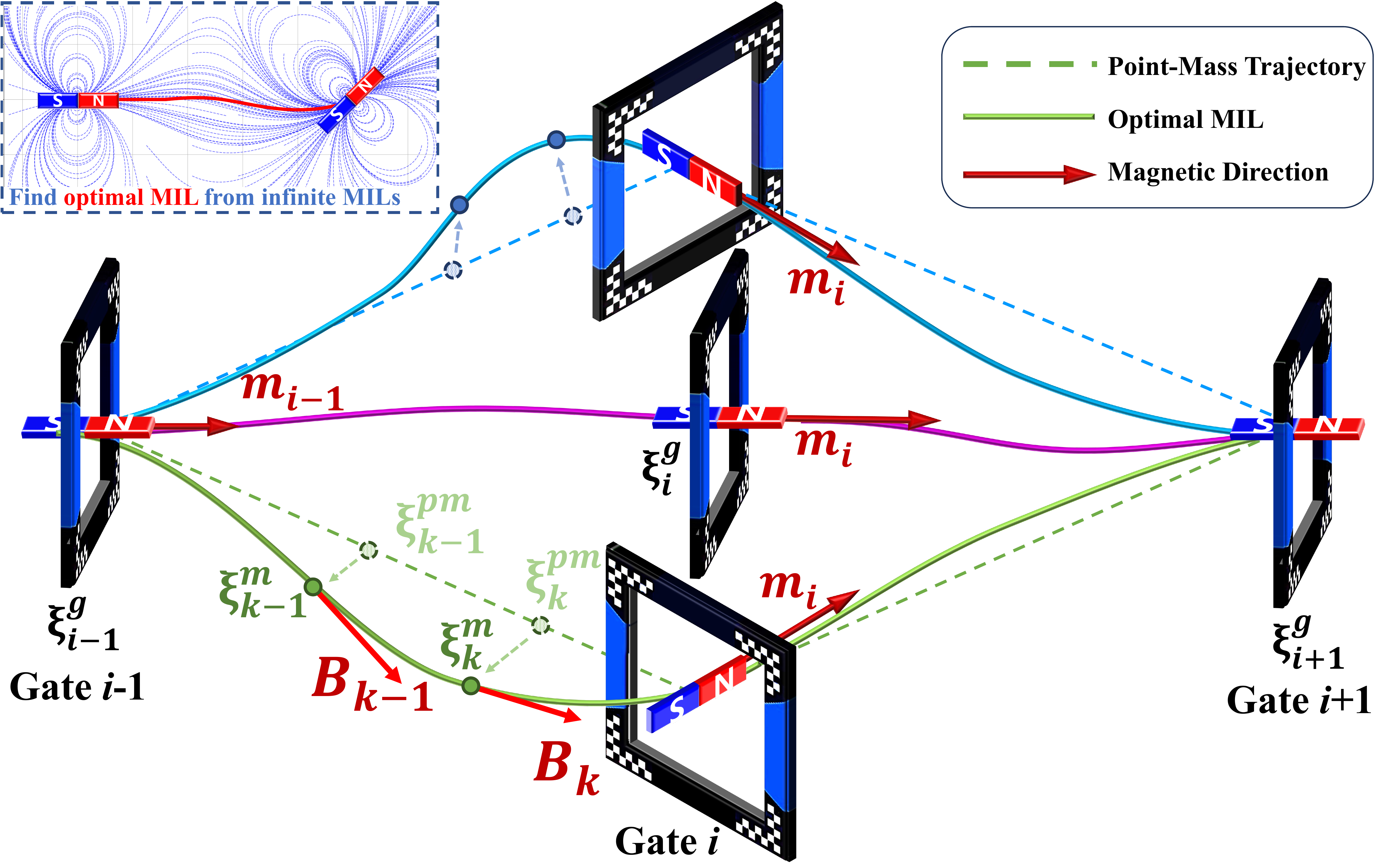}
    \caption{We use the magnet model to find the optimal MIL (solid curve) which (a) connects two magnets/gates, (b) is perpendicular to the magnets/gates at both ends (c) is the closest MIL to the point-mass reference trajectory (dashed curve) among the infinite MILs.
        }
    \label{fig:mag_gate}
    \vspace{-1em}
\end{figure}

We find that a Magnetic Induction Line (MIL) can satisfy these three requirements because (a) it can connect two magnets, (b) its start and end segments are perpendicular to the magnets, and (c) there are infinite MILs between two magnets, allowing us to find the one closest to the originally designed point-mass trajectory. Inspired by the properties of MILs, we model each gate $\boldsymbol{\xi}_i^{g}$ as a 3D magnet with a dipole moment $\boldsymbol{m}_i$ aligned with its direction, as shown in Fig. \ref{fig:mag_gate}. We then establish an optimization problem to find the optimal MIL that is perpendicular to both gates and deviates the least from the original optimal point-mass trajectory.

It is assumed that the MILs are only affected by the two adjacent magnets/gates. Then the magnetic field
$\boldsymbol{\mathcal{B}}_k$ at the $k^{th}$ point $\boldsymbol{\xi}_k^m$ on the MIL connecting two magnets can be calculated by
\begin{equation}
    \boldsymbol{\mathcal{B}}_k = \sum_{i=0}^1{\frac{\mu_0 }{4\pi}\left(\frac{3(\boldsymbol{m}_i\cdot \boldsymbol{r}_i)\boldsymbol{r}_i}{(\|\boldsymbol{r}_i\|_2^2)^5}-\frac{\boldsymbol{m}_i}{(\|\boldsymbol{r}_i\|_2^2)^3}\right)},
\end{equation}
where $\boldsymbol{r}_i=\boldsymbol{\xi}_k^m -\boldsymbol{ \xi}_i^{g}$ is the distance from the $k^{th}$ point on the MIL to the $i^{th}$ magnet/gate and $\mu_0$ is the vacuum permeability constant.
The optimization problem can be written as
\begin{equation}
    \begin{aligned}
  &\underset{\boldsymbol{\xi}_k^m}{\
  \min}\quad{\mathcal{L}_{mag}}&&=\sum_{k=1}^{N_m} \left\|\boldsymbol{\xi}_k^m - \boldsymbol{\xi}_k^{pm}\right\|_2^2 +  \left\|\boldsymbol{\xi}_k^m - \boldsymbol{\xi}_{k-1}^m\right\|_2^2\\
  & \text{subject to} & & \frac{\|\boldsymbol{\xi}_k^m -\boldsymbol{\xi}_{k-1}^m\|}{\|\boldsymbol{\xi}_k^m - \boldsymbol{\xi}_{k-1}^m\|_2} = \boldsymbol{\mathcal{B}}_k  , \\
  &&& \boldsymbol{\mathcal{B}}_k \cdot  \boldsymbol{\mathcal{B}}_{k-1} > 0.99,
\end{aligned}
\label{eq:mag_ocp}
\end{equation}
where $\boldsymbol{\xi}_k^m$ is the position of the $k^{th}$ point on the MIL and $\boldsymbol{\xi}_k^{pm}$ is the corresponding point-mass' position, which is also used as the initial value for solving the optimization problem. The first term minimizes the distance between the MIL and point-mass trajectory and the second term minimizes the distance between the adjacent points. The second term is used to avoid curl when close to the magnet as shown in the upper left corner of Fig.\ref{fig:mag_gate} where part of the MILs start to bent when close to the magnets. The optimization problem (\ref{eq:mag_ocp}) tries to find the closest MIL to the point-mass reference trajectories while being perpendicular to the two magnets/gates.

\section{Pairwise Model Predictive Control with Collision and Gate Constraints}
\label{sec:PMPC}
Although the strategy proposed in the previous section can guide a point-mass through the gates aggressively, it does not guarantee dynamic feasibility or collision avoidance between the two quadrotors. In this section, we introduce an augmented optimization problem called Pairwise Model Predictive Control (PMPC), which incorporates the dynamics of \textit{both} quadrotors as they navigate through all the gates. This approach can be executed online and inherently avoids collisions without the need for additional checks.

The quadrotors' dynamics model we use is the same with \cite{foehn2021time}. For the reader's convenience, we list the dynamics model here. It should be noted that we add the left subscript $j$ to the variables/states to denote that they belong to the $j^{th}$ quadrotor. 
\begin{align}
\prescript{}{j}{\dot{\mathbf{x}}} = \mathbf{f}_{dyn}(\prescript{}{j}{\mathbf{x}},\prescript{}{j}{\mathbf{u}}) =
\begin{cases}
\prescript{}{j}{\mathbf{v}} \\
\mathbf{g}+\frac{1}{\prescript{}{j}{m}}\mathbf{R}(\prescript{}{j}{\mathbf{q}}){\prescript{}{j}{\mathbf{T}}} \\
\frac{1}{2}\Lambda(\prescript{}{j}{\mathbf{q}})
\begin{bmatrix}
0 \\ \prescript{}{j}{\boldsymbol{\omega}} 
\end{bmatrix}
\end{cases},
\label{equ:simulation model}
\end{align}
where 
\begin{align*}
    {\prescript{}{j}{\mathbf{T}}} = \begin{bmatrix}
        0 \\  0 \\ \sum \prescript{}{j}{{T}_s}
    \end{bmatrix}
\end{align*}
is the thrust vector, $\prescript{}{j}{{T}_s}$ is the thrust of the rotors. In the equations above $\prescript{}{j}{\mathbf{x}}=[\prescript{}{j}{\boldsymbol{\xi}}, \prescript{}{j}{\mathbf{v}}, \prescript{}{j}{\mathbf{q}}]$ are position, velocity, quaternions and $\prescript{}{j}{\boldsymbol{\omega}}$ is angular velocity of the $j^{th}$ quadrotor, respectively. $\mathbf{R}(\prescript{}{j}{\mathbf{q}})$ is the rotation matrix.

\subsection{Trajectory Tracking for Point-Mass Reference Trajectory}
\label{sec:pathfollow}
In a typical MPC framework, the cost function minimizes the summed state error between the optimized variables and the reference states within the predictive horizons. Thus, the first term of the PMPC is to track the point-mass reference trajectories
\begin{align}
\prescript{}{j}{\mathcal{J}}_{track}&=&&\sum_{k=0}^N \left\|\prescript{}{j}{\mathbf{x}}_k-\prescript{}{j}{\mathbf{x}}_{ref}\right\|_{\boldsymbol{Q}_t}^2 + \left\|\prescript{}{j}{\mathbf{x}}_N-\prescript{}{j}{\mathbf{x}}_{N,ref}\right\|_{\boldsymbol{Q}_t}^2.\\
\textit{s.t.}&&&\mathbf{x}_{k+1}=f(\mathbf{x}_k, \mathbf{u}_k),\\
&&&\mathbf{x}_0=\mathbf{x}_{init},
\label{equ:equ:waypoint + velocity constraints}
\end{align}
where $\boldsymbol{Q}_t = diag(Q_{\boldsymbol{\xi}}, Q_{\mathbf{v}}, Q_{\mathbf{q}})$ is weighting matrix. The point-mass reference trajectory only has position and heading states, which do not contain the tilt (roll, pitch) state references. To compute the attitude error, we represent the quaternion error separately as  $\Tilde{\mathbf{q}}_{e,tilt}$ and $\Tilde{\mathbf{q}}_{e,yaw}$ as described in \cite{8556372,9794477}.
We give different weights to these two terms as
$Q_{\mathbf{q}}^{yaw} > Q_{\mathbf{q}}^{tilt}$, thus the heading reference can be tracked accurately while the roll and pitch are automatically optimized.

The cost function above steers the quadrotor to track the point-mass reference trajectories as accurately as possible. To push the drone to approach the next gate as fast as possible, we add a racing term to minimize the distance to the gate 
\begin{equation}
    \prescript{}{j}{\mathcal{J}}_{racing}=\sum_{k=0}^N \left\|\prescript{}{j}{\boldsymbol{\xi}}_k-{\boldsymbol{\xi}}^{g}\right\|_{Q_r}^2.
\end{equation}
where $\boldsymbol{\xi}^{g}$ is the position of the next gate. $\mathcal{J}_{track}$ and $\mathcal{J}_{racing}$ play different roles when the quadrotors are approaching the gates. When the quadrotor is far from the gate, we want it to approach the gate as fast as possible. Thus, the racing term should have a larger weight. While the quadrotor is close to the waypoint, we want it to fly through the gate accurately. Thus, the weight of the tracking term should be increased. So, a dynamic weighting scheme should be employed to adjust the weights of $\mathcal{J}_{track}$ and $\mathcal{J}_{racing}$ during the flight. In this paper, we use sigmoid curves to generate the weights for the two terms during the flight.

\subsection{Passing Gate using Optimal MIL as Spatial Reference}
\label{sec:safety}
\begin{figure}[hbt]
    \centering
\includegraphics[width=0.48\textwidth,trim={10cm 26cm 20cm 0cm}, clip]{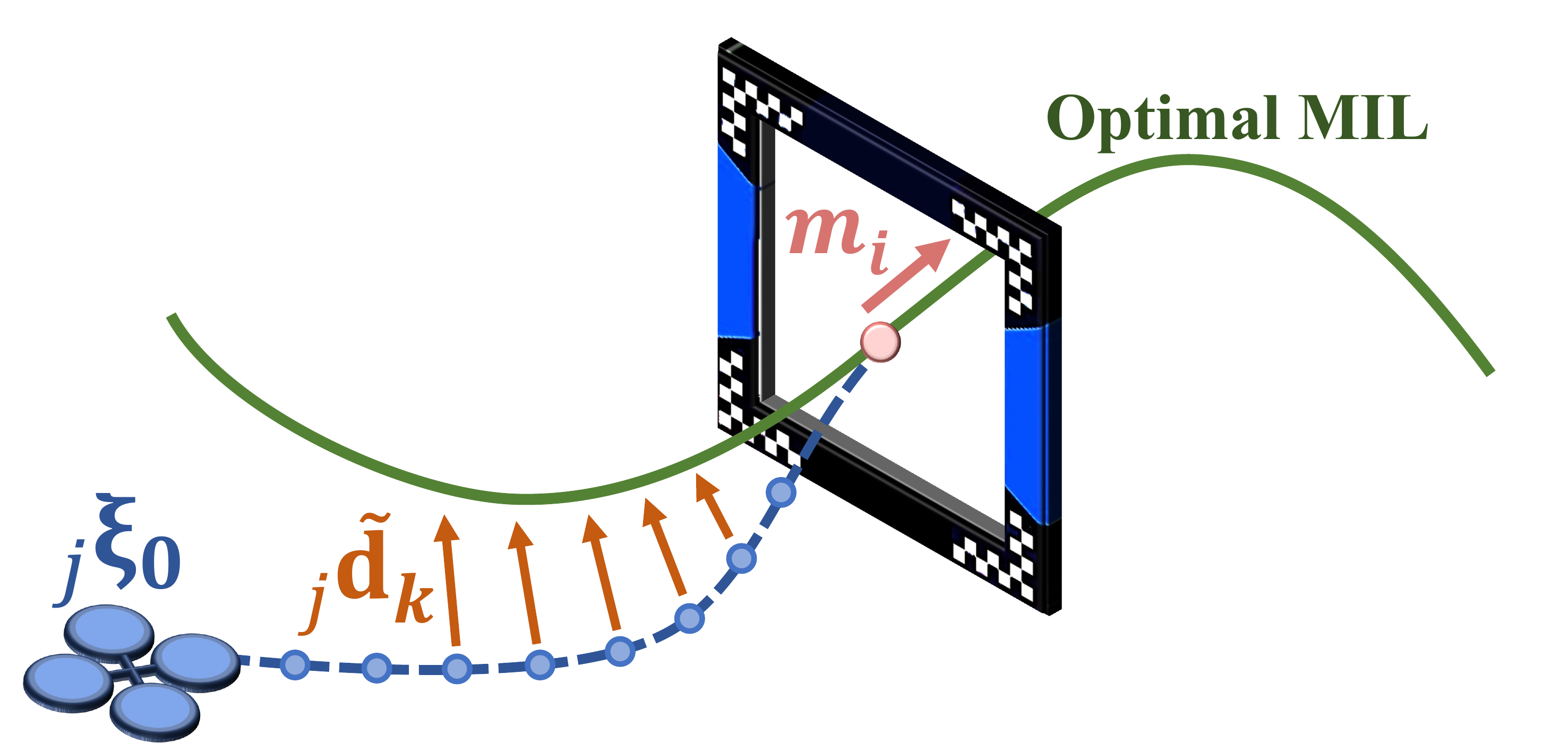}
    \caption{The MIL is used as a spatial reference to attract the quadrotor fly near the center and perpendicular to the gate.}
    \label{fig:mag_cost}
\vspace{-1em}
\end{figure}

As stated in Sec \ref{sec:mag_field}, when the quadrotor is close to the gate, the MILs should start to attract the quadrotors to guide it to fly through the gate vertically as shown in Fig. \ref{fig:mag_cost}.

Thus, we add the cost function when the distance between the quadrotor and the gate is within a pre-defined threshold.
\begin{equation}
\begin{aligned}
    \prescript{}{j}{\mathcal{J}}_{mil}  &= \sum_{k=0}^{N}\left\|\prescript{}{j}{\Tilde{\mathbf{d}}}_{k}\right\|_{Q_m}^2\\
    \prescript{}{j}{\mathbf{d}}_{k} &= (\prescript{}{j}{\boldsymbol{\xi}}_k - \prescript{}{j}{\boldsymbol{\xi}}_{mil})\\
    \prescript{}{j}{\Tilde{\mathbf{d}}}_{k} &=\prescript{}{j}{\mathbf{d}}_{k} - ( \prescript{}{j}{\mathbf{d}}_{k}\cdot \prescript{}{j}{\mathbf{m}}_i)\prescript{}{j}{\mathbf{m}}_i
    \label{eq:gate_object_mpc}
\end{aligned},
\end{equation}
where $ \prescript{}{j}{\boldsymbol{\xi}}_{mil}$ is the corresponding point on the MIL and $ \prescript{}{j}{\mathbf{m}}_i$ is the pre-defined gate passing direction. The cost function (\ref{eq:gate_object_mpc}) minimizes the distance in the direction perpendicular to $m_i$, which does not affect the flight along the direction of $m_i$.
\subsection{Predictive Collision Avoidance between Two Quadrotors}
\label{sec:collisonavoid}
The cost function mentioned above primarily focuses on single-drone flights, which do not account for interactions between two players. To address this, we incorporate collision constraints into the optimization problem to ensure that the quadrotors do not collide with each other. Rather than treating the other drone's trajectory as a constant during optimization, we address collision avoidance predictively. Each quadrotor simultaneously computes future trajectories not only for itself but also for the other quadrotor by solving the Pairwise MPC. This allows the future trajectories of both drones to be dynamically optimized, rather than being treated as a static obstacle.

Furthermore, to make the solving process easier, we only consider the collision-free constraints in part of the prediction horizon instead of the whole prediction horizon $N$. As shown in Fig. \ref{fig:main_figure}-(b) where the time-optimal trajectories of two point-mass quadrotors collide at time $t_{c,1}$ for the first time and then collide at $t_{c,2}$, in our cost function, we only evaluate the distance between two quadrotors before the first collision by
\begin{align}
   \mathcal{J}_{col}= \sum_{k=t_0}^{t_{c}}\left\|\mathbf{E}(\prescript{}{1}{\boldsymbol{\xi}_{k}}-\prescript{}{2}{}{\boldsymbol{\xi}_{k}})\right\|_{Q_c}^2,
   \label{eq:collison_objective}
\end{align}
where $\mathbf{E}$ is the matrix for relieving downwash risk \cite{zhou2021ego}, $t_c = t_{c,1}$ is the time of the first collision between the two point-mass' trajectories. This term tries to divide two quadrotors' trajectories from $t_0$ to $t_{c}$. This cost function maximizes the distance before the first collision, which is a \textit{predictive collision free} fashion and works well in high speed and same direction flights. It should be noted that this term can only guarantee that the two quadrotors do not collide before $t_{c}$. They still have a chance of collision after $t_{c}$. However, due to the high efficiency of the proposed method, the optimization problem can be solved online with high frequency ($200$Hz in our case). Thus, during the flight, the optimization loop keeps running so that the quadrotors are always guaranteed to fly without collision in the upcoming time horizon.

\subsection{Optimization Problem Formulation}
To conclude, the full PMPC can be expressed as
\begin{equation}
    \begin{aligned}
  & \underset{\mathbf{u}}{\min}\quad\mathcal{J}_{PMPC}&&=\prescript{}{1}{\mathcal{J}} + \prescript{}{2}{\mathcal{J}} - \mathcal{J}_{col}\\
  & \text{subject to} & & \mathbf{x}_0 = \mathbf{x}_{init} , \\
  &&&\mathbf{x}_{k+1} = f(\mathbf{x}_k, \mathbf{u}_k)  ,\\
  &&& \mathbf{u}_{\min}\le \mathbf{u}\le \mathbf{u}_{\max},
\end{aligned}
\label{eq:full_ocp}
\end{equation}
where $\prescript{}{j}{\mathcal{J}}=\prescript{}{j}{\mathcal{J}}_{track} + \prescript{}{j}{\mathcal{J}}_{racing}+\prescript{}{j}{\mathcal{J}}_{mil}$, $j=\{1, 2\}$, $\mathbf {x}$ and $\mathbf{u}$ are vectors of the states and inputs of the two quadrotors. The first two terms ensure that the quadrotors can pass the pre-defined waypoints with the minimum time and the third term guarantees that they do not collide with each other in a short time horizon. The constraints ensure that the optimal trajectories start from the quadrotors' current states and are dynamically feasible. With the optimal point-mass trajectories as the initial value, this optimization problem can be solved online with high efficiency. During the flight, the PMPC can run online to control the quadrotors to fly through the waypoints with time optimality without collision with each other.
\section{Experiments and Analysis}
\subsection{Simulation Results and Analysis}
In this section, we design two simulation experiments to verify the effectiveness of the proposed method. The quadrotor model we used in the simulation is (\ref{equ:simulation model}). The parameters used in the PMPC are listed in Table \ref{tab:param}.
\begin{table}[h]
\caption{The parameters used in simulation}
\vspace{-1.5em}
\label{tab:param}
\begin{center}
\begin{tabular}{c|c|c|c}
\hline
parameters & value/range & parameters & value/range \\  \hline
$N$ & $20$ & $T_{s}[N]$ & $[1,40]$ \\  \hline
$\omega_{z}[rad]$ & $[-1,1]$ & $\omega_{x,y}[rad]$ & $[-5,5]$\\ \hline
$\delta_{col}[m]$ & $0.5$ & $\mathbf{E}$ & $diag(1,1,1/3)$\\ \hline
\end{tabular}
\end{center}
\label{tab:parameters}
\vspace{-1.5em}
\end{table}        
The optimization problem (\ref{eq:full_ocp}) is solved using acados\footnote{https://github.com/acados/acados/} as the code generation tool based on \textit{Agilicious} \cite{doi:10.1126/scirobotics.abl6259} and hpipm as the solver. The solving time is about $1 ms$ on a laptop and $4 ms$ on an embedded computer Coolpi 4B \footnote{https://cool-pi.com/category/6/pi-4b} with a RAM of 16G, which is used as the onboard computer in our real world experiments.

\begin{figure}[h]
  \centering
      \centering
\includegraphics[width=0.48\textwidth,trim={1cm 0cm 2cm 0}, clip]{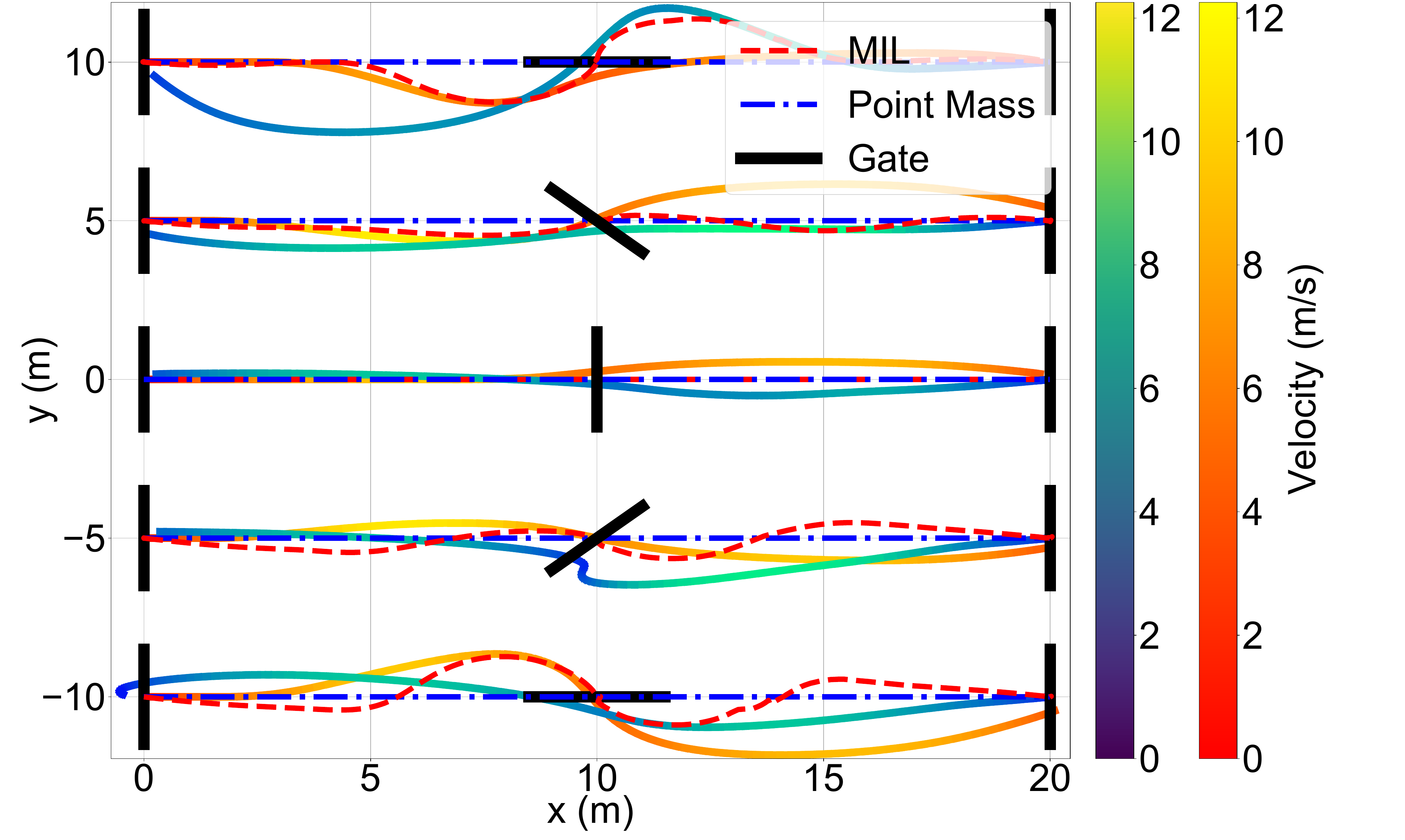}
\vspace{-2em}
  \caption{Trajectories of two drones hover to hover transition flight, the second gate has five different orientations. PMPC is collision free and the optimal MIL (dashed red curve) can ensure passing gates with different orientation.}
  \label{fig: mag_gate_two}
\end{figure}

We first designed a simple position-switching experiment to demonstrate the feasibility of the proposed collision avoidance strategy and the effectiveness of the Magnetic Induction Lines (MILs) when the quadrotors approach the gate. In this experiment, two quadrotors initially hover at $[0, 0, 2]$ and $[20, 0, 2]$, respectively. Their target positions are each other's starting points. A gate is positioned at $[10, 0, 2]$ with different orientations for the quadrotors to fly through. The results are shown in Fig.~\ref{fig: mag_gate_two}. The blue dashed curves represent the point-mass reference trajectories at the first step of the simulation. It can be observed that these reference trajectories, which do not account for collision avoidance or gate-passing directions, are nearly straight lines from the current position to the target. The red dashed curves represent the designed MILs that guide the quadrotors to fly through the gates in specific orientations. The PMPC successfully steered the two quadrotors through the middle gates, allowing them to reach their targets without collisions, as illustrated by the red and green trajectories. Notably, one quadrotor follows a relatively straight trajectory, while the other undertakes more conflict avoidance maneuvers, resulting in a more curved trajectory to minimize the total arrival time.

\begin{figure}[h]
  \centering
      \centering
      \includegraphics[width=0.48\textwidth,trim={0 0cm 4cm 0}, clip]{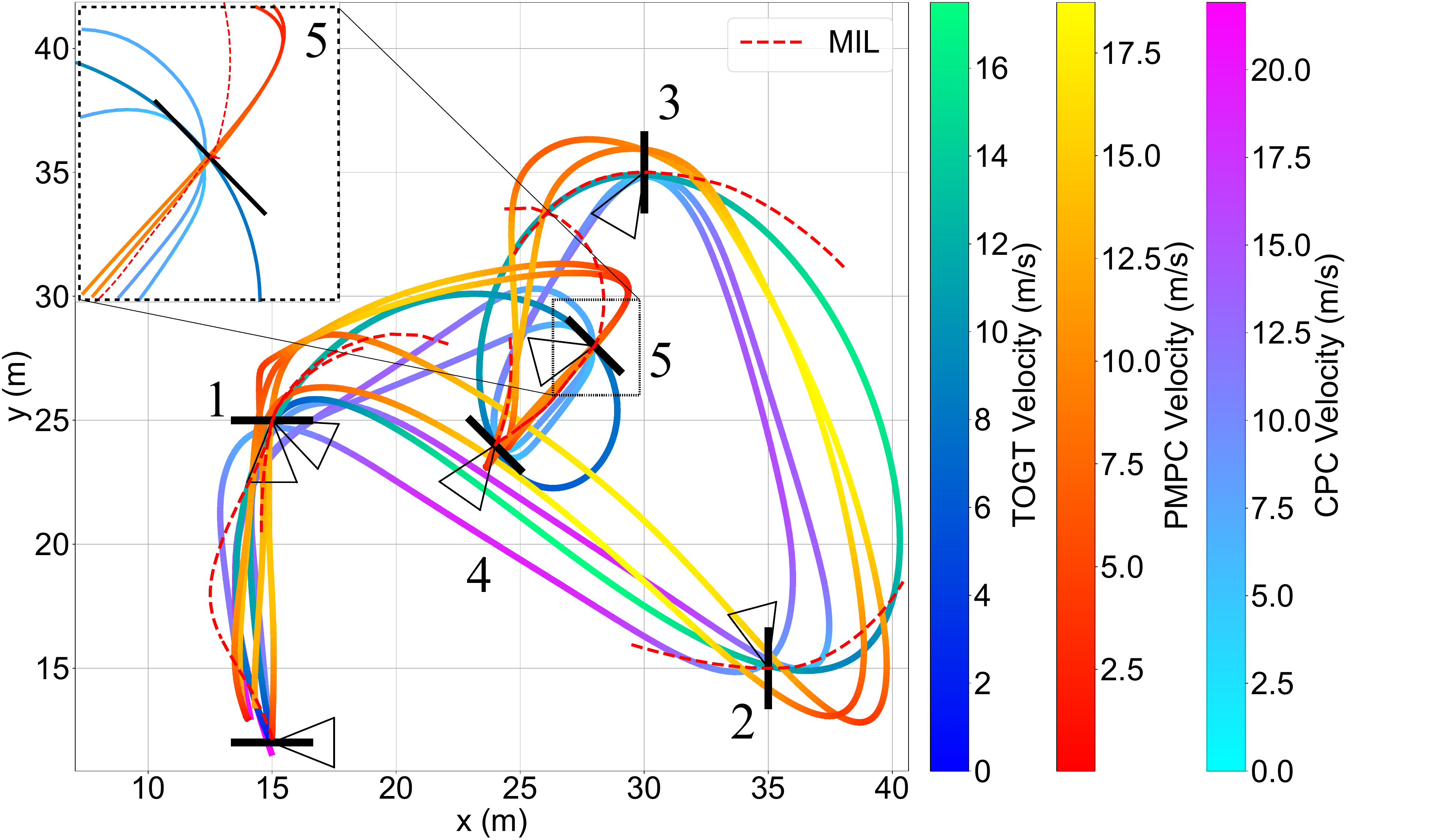}
      \vspace{-1em}
  \caption{Trajectories of quadrotors with the PMPC, TOGT and swarm-CPC. The black triangle in each waypoint denotes the optimal heading. The red dashed curve is the optimal MIL used when approaching gates. For simplicity, we only draw the MIL near the gate. TOGT and swarm-CPC can not guarantee to pass the $5^{th}$ gate precisely where PMPC can pass.}
  \label{fig: simulation_track}
  \vspace{-1em}
\end{figure}

In the second simulation experiment, we aimed to emulate a drone racing scenario using the same track as in our previous work \cite{10341844}. The experiment compared the performance of the PMPC method against the TOGT \cite{10610148} and our previous work Swarm-CPC.\cite{10341844}. The drone parameters used in the simulation are listed in Table \ref{tab:parameters}.
The TOGT method, which focuses on guiding a single drone, serves as a benchmark for MINCO-based trajectory. In contrast, both the PMPC and Swarm-CPC methods were tested with two quadrotors. The simulation results are depicted in Fig.\ref{fig: simulation_track} and summarized in Table \ref{tab:compare}.
In this experiment, the PMPC method successfully guided the two quadrotors through all the gates in a predefined sequence. As shown in Table \ref{tab:compare}, the PMPC method's arrival time is slower than that of the Swarm-CPC but faster than that of the TOGT. The Swarm-CPC achieves the fastest arrival time because it does not require the drones to end in a hover state, allowing them to rush at maximum speed. However, both the PMPC and TOGT methods require the drones to end in a hover state, which slightly increases their completion time.
Despite this, the PMPC method achieves real-time performance, effectively guiding the quadrotors through the gates in the required directions while accounting for necessary heading adjustments for better gate perception. In contrast, the Swarm-CPC method, although optimal in trajectory, requires hours to compute, making it less practical for real-time applications.

\begin{table}[h]
\caption{Comparison with the benchmark}
\vspace{-1.5em}
\label{table_example}
\begin{center}
\begin{tabular}{c|c|c|c}
\hline
&swarm-CPC & TOGT &PMPC\\\hline
solving time ($s$) &23.25k &0.018 &online\\\hline
top speed ($m/s$)&22 &17.5 &18.7\\\hline
flying time ($s$) & 10.6&12.56 &12.51\\\hline
min distance ($m$)& 0.55 & - & 0.68\\\hline
\end{tabular}
\end{center}
\label{tab:compare}
\vspace{-2em}
\end{table}
\subsection{Real‐world Experiments}
\begin{figure}[htb]
    \centering
    \includegraphics[width=0.48\textwidth,trim={0cm  0cm  0cm 0cm}, clip]{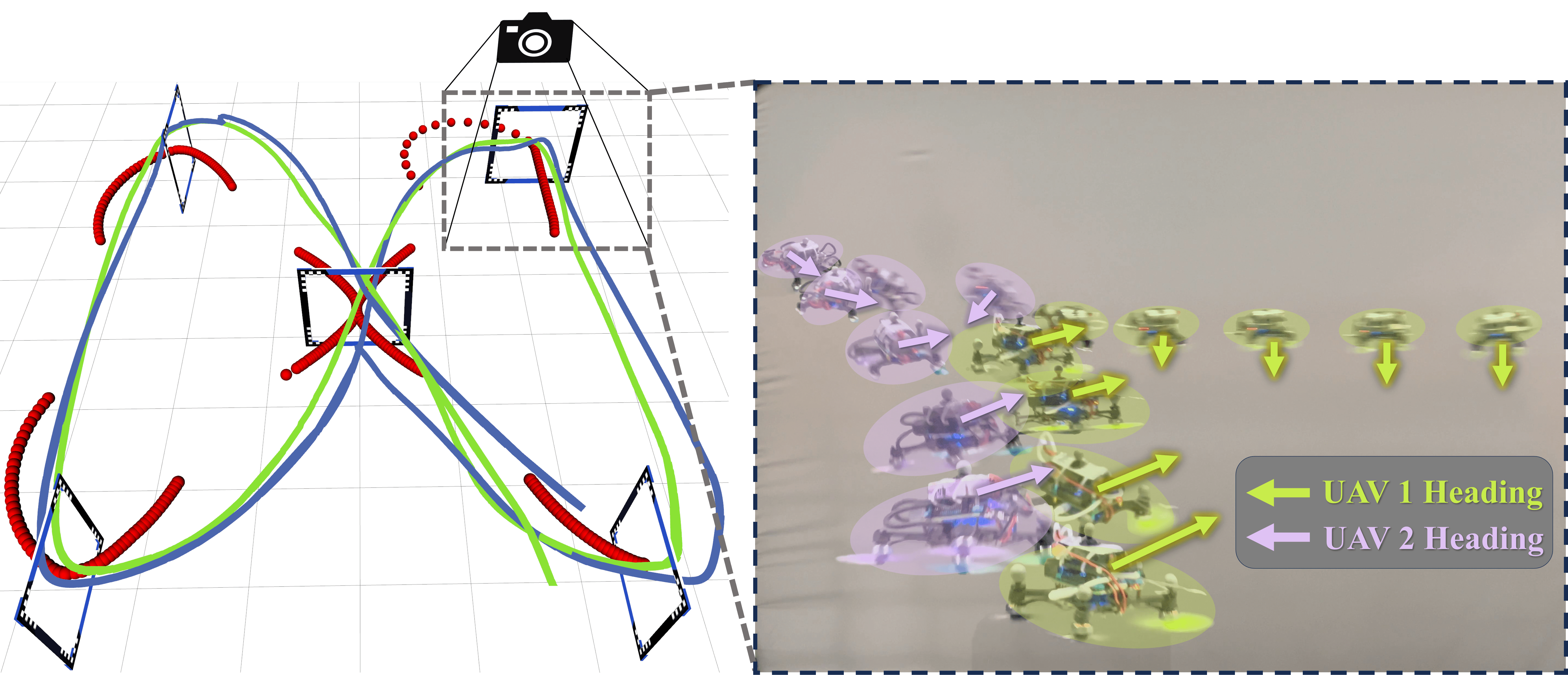}
    \caption{Left: The trajectories of the two quadrotors in the racing track, red dashed curve is the MILs used near the gate. Right: Side view of the drones. The heading of two drones points to next gate.} 
    \label{fig: real_track}
    \vspace{-1em}
\end{figure}

We use two self-developed $300g$ quadrotors as our flying platforms as shown in \cite{zhou2024imitation}. The flight space is a $5m \times 4m \times 2m$ free space as shown in Fig. \ref{fig:two quadrotor racing photo}.
We design an $8$ shape racing track which has $7$ waypoints with two different passing direction as shown in Fig. \ref{fig: real_track}.
In this experiment, two quadrotors start from their hover positions and pass through the same gates with a top speed of $6.1 m/s$ without collision. The MIL can deal with various gate orientations. The heading during the flight always points to the next gate as shown in right part of Fig.\ref{fig: real_track}. From the experiments, we can see that the proposed method, the PMPC, performs well and can guide the quadrotors online to fly through the waypoints with their maximum maneuverability without collision. 
    \vspace{-1em}
\section{Conclusion}
In this paper, we presented a systematic method called Pairwise Model Predictive Control (PMPC) for enabling two autonomous racing drones to fly through waypoints with varying orientations in a predefined sequence. The method begins with a velocity-heading search using a simple point-mass model to generate global time-optimal trajectories as references. Subsequently, based on the spatial configuration of the track, an optimal Magnetic Induction Line (MIL) is generated to serve as a spatial reference when approaching the gate. An optimization problem is then formulated to account for the quadrotors' dynamics and ensure predictive collision avoidance during flight. Compared to the benchmark method, PMPC demonstrates high computational efficiency. Both simulation and real-world experiments confirm that the proposed method effectively enables collision avoidance for two racing drones.
Several avenues for future research have been identified. For instance, we plan to extend our work to accommodate a larger number of quadrotors, thereby facilitating its application to a swarm of drones. Additionally, we aim to develop methods for generating MILs online to handle dynamic gates. Furthermore, we intend to explore the inclusion of tilt motion in perception-aware tasks.
\newpage
\bibliography{reference}
\bibliographystyle{ieeetr}
\end{document}